\documentclass[10pt,twocolumn,letterpaper]{article}

\usepackage{cvpr}
\usepackage{times}
\usepackage{epsfig}
\usepackage{graphicx}
\usepackage{amsmath}
\usepackage{amssymb}
\usepackage{amsmath}
\usepackage{float}

\usepackage[pagebackref=true,breaklinks=true,letterpaper=true,colorlinks,bookmarks=false]{hyperref}

\cvprfinalcopy 

\def\cvprPaperID{4092} 

\ifcvprfinal\fi
\begin{document}

\title{Fast Recurrent Fully Convolutional Networks for Direct Perception in Autonomous Driving}

\author{Eric (Yiqi) Hou\\
Berkeley Deep Drive\\
109 McLaughlin Hall\\
Berkeley, CA 94720-1720\\
{\tt\small eric.hou@berkeley.edu}
\and
Sascha Hornauer\\
University of California, Berkeley\\
(ICSI) 1947 Center St\\
Berkeley, CA 94704\\
{\tt\small saschaho@icsi.berkeley.edu}
\and
Karl Zipser\\
Berkeley Deep Drive\\
109 McLaughlin Hall\\
Berkeley, CA 94704\\
{\tt\small karlzipser@berkeley.edu}
}

\maketitle

\begin{abstract}
 Deep convolutional neural networks (CNNs) have been shown to perform extremely well at a variety of tasks including subtasks of autonomous driving such as image segmentation and object classification. However, networks designed for these tasks typically require vast quantities of training data and long training periods to converge. We investigate the design rationale behind end-to-end driving network designs by proposing and comparing three small and computationally inexpensive deep end-to-end neural network models that generate driving control signals directly from input images. In contrast to prior work that segments the autonomous driving task, our models take on a novel approach to the autonomous driving problem by utilizing deep and thin Fully Convolutional Nets (FCNs) with recurrent neural nets and low parameter counts to tackle a complex end-to-end regression task predicting both steering and acceleration commands. In addition, we include layers optimized for classification to allow the networks to implicitly learn image semantics. We show that the resulting networks use 3x fewer parameters than the most recent comparable end-to-end driving network \cite{Nvidia} and 500x fewer parameters than the AlexNet variations and converge both faster and to lower losses while maintaining robustness against overfitting. 
\end{abstract}

\section{Introduction}
\label{sec:introduction}
The autonomous driving task is an extremely broad and challenging problem. Consequences of failure, large input spaces, and relative complexity of control and navigation through those highly variable spaces have caused the vast majority of research to focus on subtasks of the problem such as lane detection \cite{DeepLane} and obstacle recognition \cite{3D-OD}, \cite{Perception}.

However, state-of-the-art models for these segments of autonomous driving are often extremely parameter-dense and require high amounts of computational power. 3D object detection \cite{3D-OD}, for instance, takes 0.36s for inference alone on a Titan X, or less than 3 inferences per second, and relies on a modified but still parameter-heavy VGG-16. While extremely accurate, with modern-day processor limitations, such models are too heavy to reach better than human reaction times of 100ms (10FPS) compared to traditional systems like LIDAR (24 FPS) \cite{LidarOD} and algorithms based on traditional machine learning (100 FPS) \cite{HOG-OD} that achieve higher FPS at the expense of lower accuracy.

\begin{figure}
\begin{center}
\includegraphics[scale=0.11]{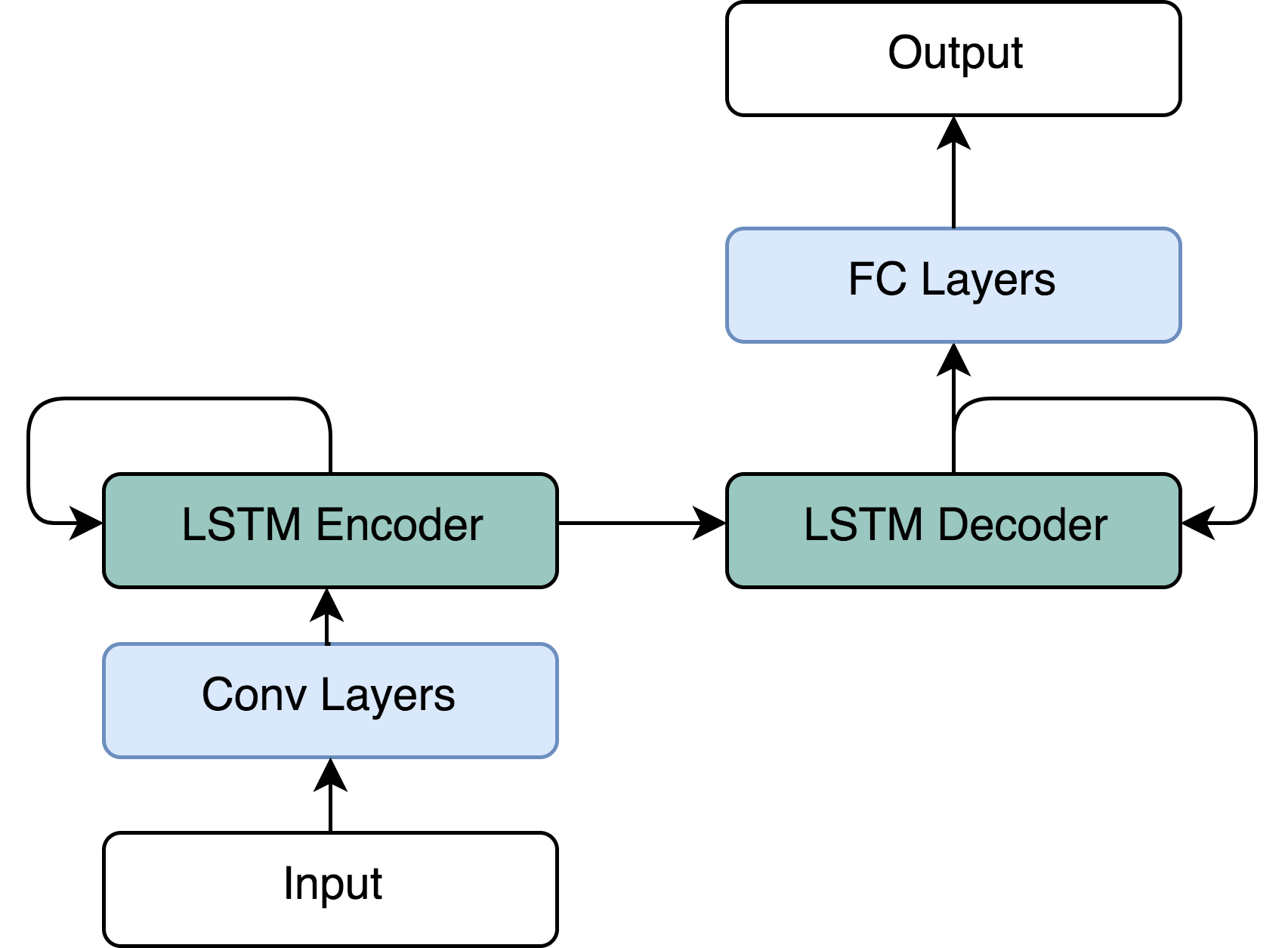}
\end{center}
\caption{F-RFCN Architecture}
\label{fig:short}
\end{figure}

As an alternative to segmented tasks, recent work has demonstrated promise in deep end-to-end architectures for direct perception in autonomous driving \cite{Nvidia}, \cite{DeepDriving}, \cite{OffroadE2E}, \cite{MTL}. Here, we define end-to-end as architectures that map directly from video input to control outputs. Some approaches rely on large datasets to reduce temporal correlation between different training examples to increase accuracy \cite{LargeE2E}. Most notably, the work by \cite{Nvidia} has demonstrated 98\% vehicle autonomy using only a standard convolutional neural network, recorded video, and steering angle from vehicles. These end-to-end systems have demonstrated the feasibility of highly autonomous end-to-end imitation learning for autonomous vehicle control with low parameter counts. We will explore more of this prior work in Section [\ref{sec:related-work}].

We expand on the end-to-end driving task by examining prior work on subtasks of autonomous driving and computer vision and using their insights to generate more informed models. We ultimately describe the fast recurrent fully-convolutional network (F-RFCN) architecture for end-to-end driving. 

These models show a few novel insights: firstly, we empirically demonstrate the ability of extremely small convolutional networks to tackle the end-to-end autonomous vehicle control problem. Our networks are able to reduce parameter size by 3x compared to the next smallest end-to-end network and converge faster than alternative models. Secondly, we show that the theoretical performance at convergence beats that of the Nvidia model. Finally, we demonstrate that these networks still learn and generalize extremely well with sparse data.

\section{Related Work}
\label{sec:related-work}
End-to-end autonomous driving for vehicles was first explored by the ALVINN system \cite{ALVINN} to exploit the emerging power of neural networks to directly output driving controls. This preliminary work, which was a DARPA-funded project over eight years at Carnegie Mellon University, utilized a shallow fully-connected network with some feedback mechanisms to output controls and experimented with dataset augmentation. This success demonstrated the feasibility of end-to-end autonomous vehicle control and laid the groundwork for training and architecture for future work.

More recently, work within supervised training for end-to-end control has led to significant improvements in imitation-based autonomous driving. A standard CNN from \cite{Nvidia} (DAVE-2) demonstrated success in highly structured environments, such as highway lane following and driving on flat terrain. However, this network demonstrated an inability to turn or change lanes, which was attributed to the lack of laterally-facing camera input.

Thus, in our search for a more sophisticated model, we look towards multi-modal and multi-task learning (MTL). MTL is a relatively new technique from \cite{MTL-original} that was introduced to the direct perception problem by \cite{BDD-MTL} in the Z2Color net, which demonstrated that modal information and multi-task learning lowers validation losses and raises autonomy across different modes. They also train on multiple tasks by asking for higher-resolution output control prediction than is actually used, with the unused time steps acting as side tasks. 

\cite{MTL} furthers the idea of modal data to train their models using vectors embedded in a latent modal space. They provide mathematical intuition for their work by showing that the model implicitly learns output controls as expressive functions of the modal data. While both MTL networks work well experimentally, they do so at a cost of model size and computational complexity.

Another line of related work utilizes recurrent neural networks (RNNs) to help train models. Specifically within the autonomous driving space, \cite{LargeE2E} utilizes recurrent fully-convolutional networks (RFCNs) based on AlexNet to predict output steering commands at each timestep given the frame at each timestep. Like the MTL and multi-modal models, the RFCN includes subtask-based training to help with accuracy of the model. Additionally, \cite{LargeE2E} shows that the FCN and CNN architectures perform at almost the same accuracy, with FCNs theoretically containing far fewer parameters.

As mentioned in Section [\ref{sec:introduction}], much related work has established feasibility of deep convolutional neural networks in autonomous vehicle control, but there is little work on practical architecting of models to optimize model size, training time, required dataset size, and stability of convergence. The SqueezeNet and SqueezeDet convolutional models from \cite{SqueezeNet} and \cite{SqueezeDet} respectively have resulted in models tackling those practical motivations. More interestingly, they have been proven to reach near state-of-the-art performance on the object classification and object detection subtasks of the autonomous vehicle problem while maintaining significantly fewer parameters.

\section{Approach}
\label{sec:approach}
We formulate the learning problem based on the prior work by Nvidia \cite{Nvidia} and work from the Berkeley Deep Drive project \cite{BDD-MTL}, \cite{LargeE2E}. We take in the past 6 video frames at 10 FPS as input into the networks and use them to predict the next 12 human-controlled steering and motor output controls at 10 FPS in an imitation learning format. Intuitively, these predicted 12 steering-motor pairs represent a planned trajectory over a few time-steps even though only one pair is used for the next control output. To approach the problem, we combine and extend both the segmented and end-to-end approach of architecting autonomous systems to create several semi-architected models that tackle the imitation learning approach to end-to-end autonomous driving. In particular, we approach the regression task of direct perception by examining existing models and paradigms that have shown promise in subtasks. 

By including network paradigms from classification \cite{SqueezeNet}, visual recognition and image semantics \cite{RFCN-Recognition},  \cite{RFCN-Semantics}, and object tracking \cite{ObjectTracking}, \cite{SqueezeDet}, we hypothesize that our model implicitly learns how to incorporate this analysis in its internal semantic structure. Drawing from classification, we follow the motivation of SqueezeNet \cite{SqueezeNet} to significantly reduce the base size of our model while retaining its expressiveness. Analogous to visual recognition and image semantics approaches \cite{RFCN-Recognition}, \cite{RFCN-Semantics}, we design an encoder-decoder network to allow the convolutional net to learn semantic representations of the input images with respect to controls: we include recurrent layers such as LSTMs used in object tracking \cite{ObjectTracking} to encode spatiotemporal data and decode output controls. As mentioned previously, the resulting networks tackle the end-to-end problem with fewer parameters, better theoretical performance, and faster convergence.

\section{Models}
\begin{figure*}
\begin{center}
\includegraphics[scale=0.11]{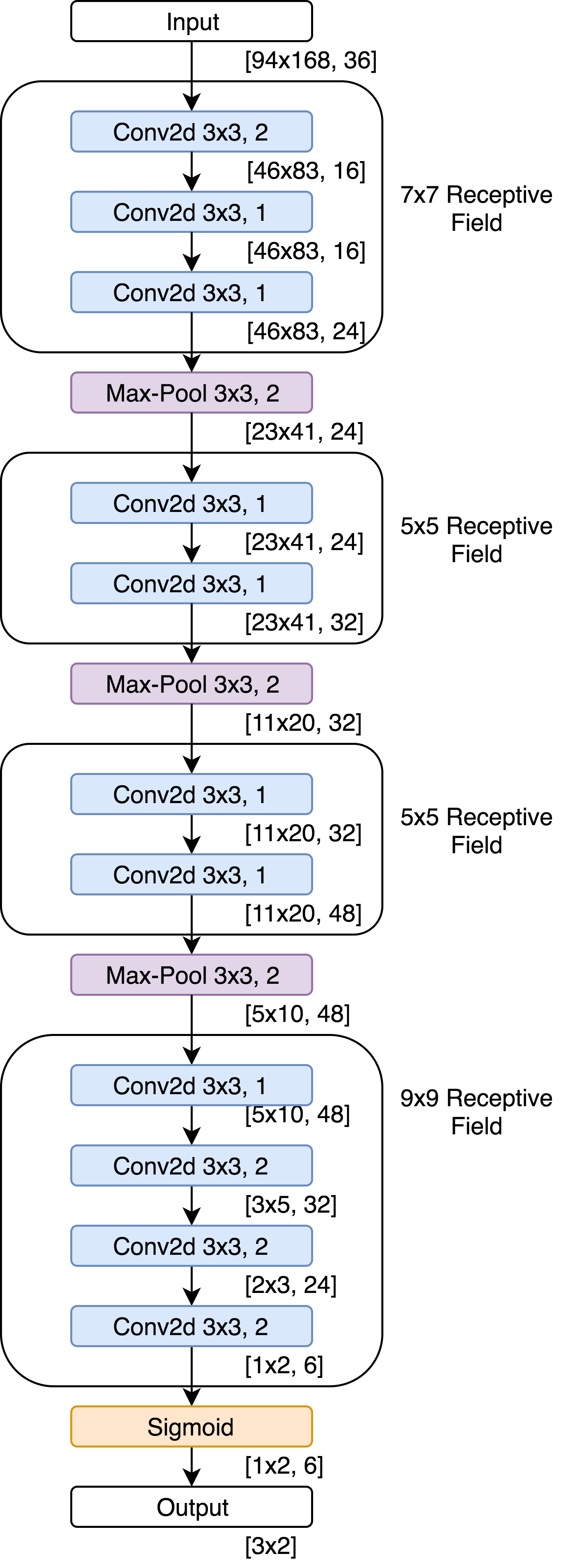}
\hspace{1cm}
\includegraphics[scale=0.11]{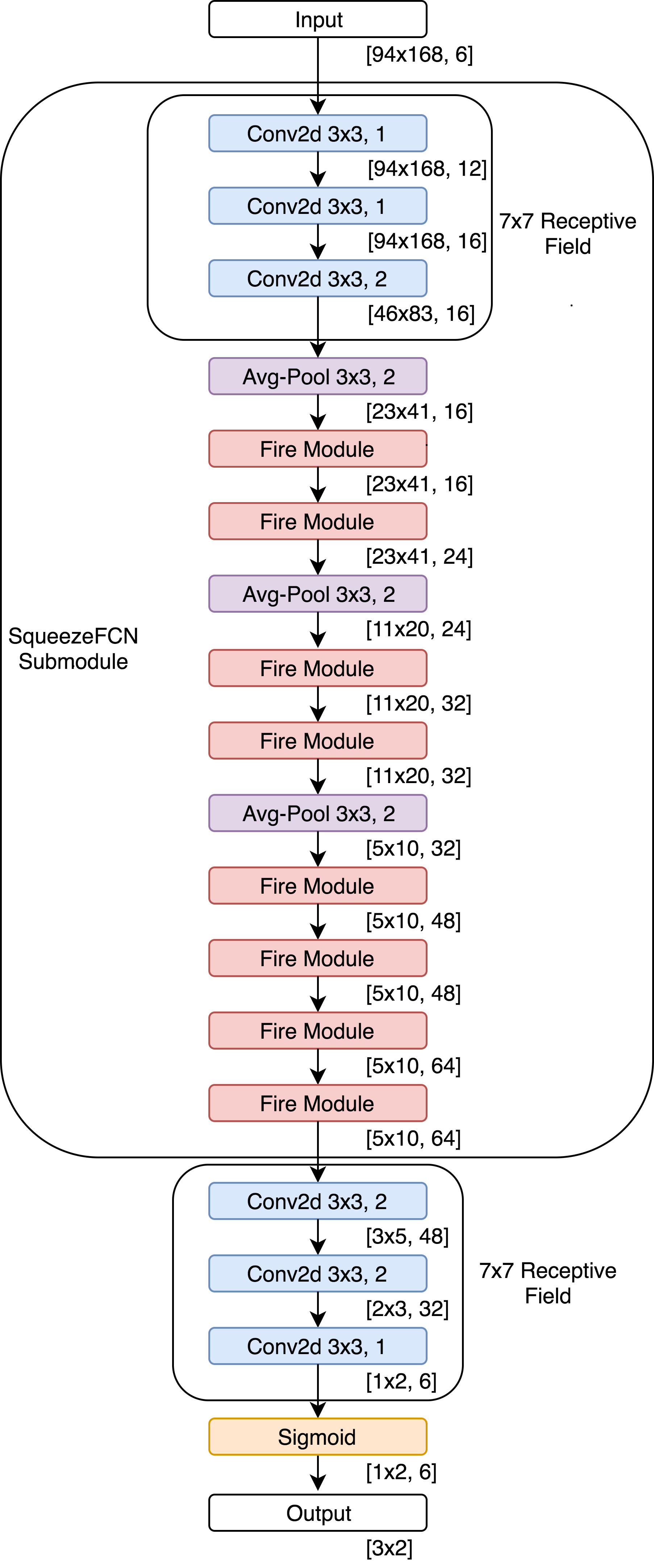}
\end{center}
\caption{The feedforward architectures. Left: FCN. Right: SqueezeFCN. We label layers with the convention [layer conv-size, stride]. We label layer inputs/outputs as [spacial size, feature depth] where n=batch size.}
\label{fig:base-model}
\end{figure*}

Motivated by \cite{LargeE2E}, we frame the general class of models we are constructing as generic driving models, which learn a driving policy given by an imitation-based dataset by predicting future steering and motor commands. We begin by describing a naive deep FCN with 11 layers, an extremely low parameter count, and simple design, as well as a classification-motivated network with a similar architecture. We also incorporate a recurrent architecture mentioned in our approach (Section [\ref{sec:approach}]) to create a larger time-unwrapped model, the F-RFCN, and later empirically demonstrate its ability to converge more smoothly and quickly due to its recurrence relations. Figure [\ref{fig:base-model}] compares the two feedforward architectures. Figure [\ref{fig:f-rfcn}] depicts the time-unwrapped encoder-decoder network (F-RFCN).

\subsection{Deep FCN Architecture}
\label{sec:FCN}
Given a series of video frames, feedforward CNNs and FCNs often append the images along the feature channel and take the resulting tensor as an input. This architecture encodes the difference between frames, which contain spatiotemporal data such as motion, through the convolutional layers into feature vectors. This allows the convolutional layers to implicitly encode said spatiotemporal data and utilize it to output a sequence of control outputs over time. While this is an effective approach, we demonstrate experimentally that by implicitly encoding temporal video input data in spatially motivated convolutional layers, the loss function of the model is much more variable and less monotonic, so the model converges in a less stable manner, making stopping criteria more difficult to determine. The FCN model, while not as robust of a model, is our smallest model with around 90,000 parameters.

For all of our networks, we use a significantly reduced number of channels for each layer. We also add in a nonlinear activation after every single linear or convolutional layer, and BatchNorm \cite{BatchNorm} after every single activation layer, which are not shown in the diagrams. Note that due to the nature of the end-to-end autonomous driving task, it is very difficult to extract which parameters are learning which subtasks, or if the subtasks are being learned at all.

\subsubsection{Spatiotemporal Encoding in FCN}

We utilize stacked 3x3 convolutions as described by \cite{DeepCNN} to allow for large receptive fields with fewer parameters. Motivated by \cite{Nvidia}, we begin by using a convolutional layer to downsample the image, and use a ReLU and max-pooling layer as a normalization factor. Effectively, we have receptive fields of 7x7, 5x5, 5x5, and 9x9, expressed by a total of 11 layers. Intuitively, this structure allows for early normalization and for later layers to conduct higher-level perception across wider areas of space.

\subsubsection{Multi-Task Learning}

For multi-task learning, we output a prediction of 2 controls over 12 time frames. We only use the 1st time frame, resulting in the last 11 frames acting as subtasks for multi-task learning. As demonstrated by \cite{BDD-MTL} and \cite{MTL}, these subtasks allow for better performance on the main task by providing more and higher-quality loss feedback. In the F-RFCN, we will demonstrate that the MTL architecture allows for significantly faster convergence and smoother energy functions.

\subsection{SqueezeFCN Architecture}

The SqueezeFCN architecture employs the same approach to multi-task learning as the FCN architecture. However, we use the motivation of \cite{SqueezeNet} and \cite{SqueezeDet} and their Fire layers for improved implicit classification and object tracking. Additionally, we swap out all of the ReLU activations for ELU activations.

\subsubsection{Spatiotemporal Encoding in SqueezeFCN}

We utilize a similar architecture to the FCN, replacing all but the first and last convolutional layers with Fire layers. We then replace the input and output convolutional layers with three 3x3 convolutional layers for the 7x7 receptive fields at the input and output of the SqueezeFCN. We choose ELU layers for all of our activations and 3x3 average pooling with a stride of 2 for our pooling layers. Our first convolutional layers are motivated as described in Section [\ref{sec:FCN}] and explained in SqueezeNet \cite{SqueezeNet}: instead of immediately introducing the image channels to classification-optimized layers, we allow the network to learn some form of normalization and preprocessing.

\subsection{F-RFCN Architecture}

\begin{figure}[h]
\begin{center}
\includegraphics[scale=0.13]{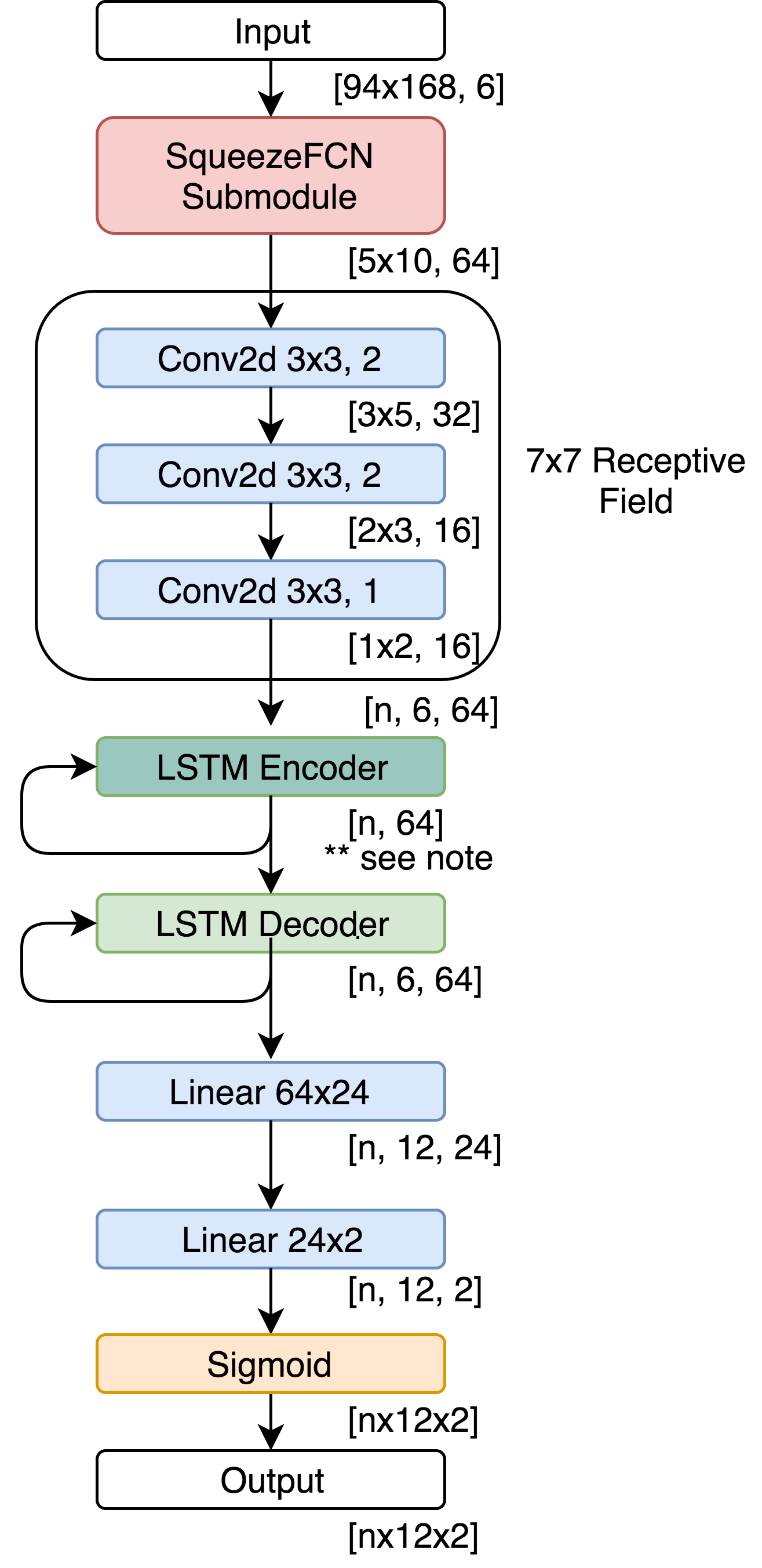}
\end{center}
\caption{F-RFCN. Note: the LSTM encoder's $h_{n}$ initializes the decoder's state $h_{0}$}
\label{fig:f-rfcn}
\end{figure}

The F-RFCN architecture utilizes the SqueezeFCN submodule, shown in Figure [\ref{fig:base-model}], as the main convolutional network. However, this network unwraps the FCN over time and applies it to each input frame, feeding the result into an LSTM. Unlike in \cite{LargeE2E}, we do not directly predict the next control signal, but we instead use an encoder-decoder structure to maximize feedback from multi-task learning. This allows the network to make a variety of internal structural changes in learning feedback. The F-RFCN model is shown in Figure [\ref{fig:f-rfcn}].

\subsubsection{Spatiotemporal Encoding in F-RFCN}

Because each input frame to the F-RFCN is passed through the SqueezeFCN submodule independently, the submodule architecture does not learn any temporal dependencies. Instead, its role in the architecture is to embed the image into a representational vector space of dimension 16 for the LSTM to further encode over time. This explicit separation of tasks allows each sub-network to focus on its strengths; the LSTM encodes and decodes sequential data, and the SqueezeFCN embeds the image into some semantic representation of the instantaneous scene. 

The LSTM decoder takes in the state encoded by the encoder and extracts it into semantic representations of information relevant to the output controls at each output frame. Then, the fully-connected layers interpret this semantic information to output control signals. This allows further separation of learning, ensuring that the LSTMs only encode and decode time-series data and do not need to learn transformations.

\subsubsection{Multi-Task Learning in F-RFCN}

Learning with LSTMs significantly changes the feedback structure of multi-task learning. Because we define the side tasks as all of the timesteps that are not used, and because this data is temporal in nature, the LSTM allows for faster convergence due to the feedback inherent in the structure of RNNs. \\

In the following we describe its gates to demonstrate the feedback of the LSTM into the FCN. We define $F_{gate}$ as the corresponding gate's function with respect to its inputs. We examine the gradient feedback provided by the LSTM equations to demonstrate mathematically why LSTMs provide for more stable training. The gates as listed below can be found in the original LSTM paper \cite{LSTM}.

\begin{align}
    \vec{f}_t &= F_{f}(\vec{h}_{t-1}, \vec{x}_{t})\\
    \vec{i}_t &= F_{i}(\vec{h}_{t-1}, \vec{x}_{t})\\
    \vec{c'}_t &= F_{c'}(\vec{h}_{t-1}, \vec{x}_{t})\\
    \vec{c}_t &= F_{c}(\vec{f}_{t},\vec{c}_{t-1},\vec{i}_{t},\vec{c'}_{t})\\
    \vec{o}_t &= F_{o}(\vec{h}_{t-1}, \vec{x}_{t})\\
    \vec{h}_t &= F_{h}(\vec{o}_t, \vec{c}_t)
\end{align}

Note that due to the recurrent relation, each $\vec{h}_t$ can be unwrapped recursively as follows:

\begin{align}
    \vec{h}_t &= F_{h}(\vec{o}_t, \vec{c}_t)\\
    &=F_{h}(F_{o}(\vec{h}_{t-1}, \vec{x}_{t}), F_{c}(\vec{f}_{t},\vec{c}_{t-1},\vec{i}_{t},\vec{c'}_{t}))\\
    &=F_{h}(F_{o}(\vec{h}_{t-1}, \vec{x}_{t}), F_{c}(F_f(\vec{h}_{t-1}, \vec{x}_t), \\
    \nonumber &\hspace{0.5cm}\vec{c}_{t-1}, F_{i}(\vec{h}_{t-1}, \vec{x}_{t}), F_{c'}(\vec{h}_{t-1}, \vec{x}_{t})))\\
    \vec{h}_t&=4T(\vec{h}_{t-1}) + T(\vec{c}_{t-1}) + 4T(\vec{x}_{t})
\end{align}

We can see that the recurrence relation for $\vec{h}_{t}$ has a branching factor of 4 on $\vec{h}_{t-1}$ and a branching factor of 1 on $\vec{c}_{t-1}$. This means at each timestep, the amount of feedback in the gradients of $\vec{h}_0$ increases by a factor of 4. The recurrence relation for $\vec{c}_{t}$ is shown below.

\begin{align}
\vec{c}_t &= F_{c}(\vec{f}_{t},\vec{c}_{t-1},\vec{i}_{t},\vec{c'}_{t})\\
\vec{c}_t&=F_{c}(F_f(\vec{h}_{t-1}, \vec{x}_t), \vec{c}_{t-1}, F_{i}(\vec{h}_{t-1}, \vec{x}_{t}), F_{c'}(\vec{h}_{t-1}, \vec{x}_{t}))\\
\vec{c}_t&=3T(\vec{h}_{t-1}) + T(\vec{c}_{t-1}) + 3T(\vec{x}_{t})
\end{align}

Here, the recurrence relation for $\vec{c}_t$ has a branching factor of 3 on $\vec{h}_{t-1}$ and a branching factor of 1 on $\vec{c}_{t-1}$. Thus, we know the recurrence relation has an upper bound that is exponential in complexity, so the amount of gradient feedback that $\vec{h}_0$ receives is exponential relative to time length of the output sequence. With multi-layer LSTMs, this complexity increases to $O(c^t{t + d \choose d})$ where $t$ is output length, $d$ is LSTM depth, and $c$ is some constant. Expanding over the encoder as well, this becomes $O(c_{in}^{t_{in}}{t_{in} + d_{in} \choose d_{in}} c_{out}^{t_{out}}{t_{out} + d_{out} \choose d_{out}})$.

Though most of the branching is highly correlated, each additional timestep in the LSTM provides an exponentially larger amount of gradient feedback, which allows for us empirically see both faster convergence and lower losses.

\section{Dataset}
We collected data using two different methods in various environments and at different times during the day, over the course of one year. Additional information in the dataset from accelerometer, gyroscope, and wheel encoder data -- which report acceleration, speed, and turning rate respectively -- may also be present. Constantly present are steering and motor commands as well as state information which label the recorded image stream from a stereo input camera.

Our first and standard method is to drive the RC model cars using a remote controller through rough environments as offroad paths through national parks, on sand-paths at a university campus and on sidewalks through different routes in the city. Additionally, we recorded data in a confined area covered by woodchips with various boundaries marked by caution tape and cones.

After the first round of data collection, we trained an initial version of the network to drive along basic paths and avoid obstacles while requiring less and less human intervention. Thus, when human intervention is required, the data that is collected is focused around scenarios that the model is not able to learn from the existing dataset. This allows for situational coverage in the data.

At the time of writing, the existing data amounted to roughly 700 GB. With basic dataset augmentation such as horizontally flipped video and control outputs, we obtained a dataset of 1.4 TB.

\section{Experiments}
Our models were trained using the Adam gradient descent method \cite{Adam} with $\alpha = 10^{-3}$,$\beta_1 = 0.9$, $\beta_2 = 0.999$, $\epsilon = 10^{-8}$, and a batch size of 32. We test two subsets of our dataset: one with all of the available training data for sampling, and one with only 10\% of the available training data for sampling to mimic the effect of having collected less data.

We validate our models two ways. First, we demonstrate theoretical improvements via loss function stability, convergence rate, and absolute loss values. Then, we test the models on real-world tracks and tasks. 

\subsection{Theoretical Improvements}

To validate how quickly our model trained, we subsampled our training set at 0.1\% of our training set per validation run. For validation, we took 20 randomly sampled, completely unseen model car driving runs, spanning a total of a few hours, and we randomly sampled 1\% of its frames to make up the validation set. 

We define the convergence rate $r$ as the geometric mean of convergence rates by epoch $r_e$ until the epoch with the minimum loss as expressed by $r_e = \dfrac{l_{e} - min_{\forall e}(l) }{l_{e-1} - min_{\forall e}(l)}$ where $l_e$ is the validation loss at epoch $e$ and $min_{\forall e}(l)$ is the minimum validation loss over all epochs. We also calculate the divergence rate $r_{div}$ using convergence rate after the minimum loss and the corresponding geometric SD factor $\sigma_r$ and $\sigma_{r, div}$ before and after convergence. To avoid skewing $\sigma_r$, which measures stability of convergence, we remove the first few epochs during its calculation where $l_e >>min_{\forall e}(l)$ so that initial training is not considered into the stability calculation. 

\subsubsection{Full Dataset Sampling}
\begin{table}[t]
\begin{center}
\begin{tabular}{c|ccccc}
\multicolumn{1}{c}{\bf}  &\multicolumn{1}{c}{\bf FCN} &\multicolumn{1}{c}{\bf SqueezeFCN} &\multicolumn{1}{c}{\bf F-RFCN} &\multicolumn{1}{c}{\bf Nvidia*}
\\ \hline \\
$r$         & \textbf{0.972} & \textbf{0.974} & \textbf{0.975} & 0.984\\
$r_{div}$   & \textbf{1.009} & 1.015 & 1.029 & 1.014\\
$\sigma_r$  & 1.018 & 1.020 & 1.124 & \textbf{1.014}\\
$\sigma_{r, div}$  & 1.037 & 1.055 & \textbf{1.024} & \textbf{1.024}\\
$min_{\forall e}(l)$ & \textbf{0.0061} & \textbf{0.0062} & \textbf{0.0064} & 0.0076\\
epochs & 102 & 102 & \textbf{89} & 96\\
Parameters & \textbf{91k} & 100k & 120k & 370k
\end{tabular}\vspace{0.2cm}
\caption{Comparison of validation loss function convergence}
\label{table:full-loss}
\end{center}
\end{table}

Table [\ref{table:full-loss}] shows an overview of our results. Firstly, we see that each of the networks beats the performance of the Nvidia model under normal sampling. We see that each of the new networks reaches a minimum loss of around $0.0062$ while Nvidia's reaches $0.0076$.  Additionally, all of the networks converge at similar rates, but the deep FCN is the most stable. The F-RFCN, as expected, converges more quickly than the other networks do, converging in 89 epochs compared to Nvidia's 96. \\

\begin{figure}
\begin{center}
\includegraphics[scale=0.35]{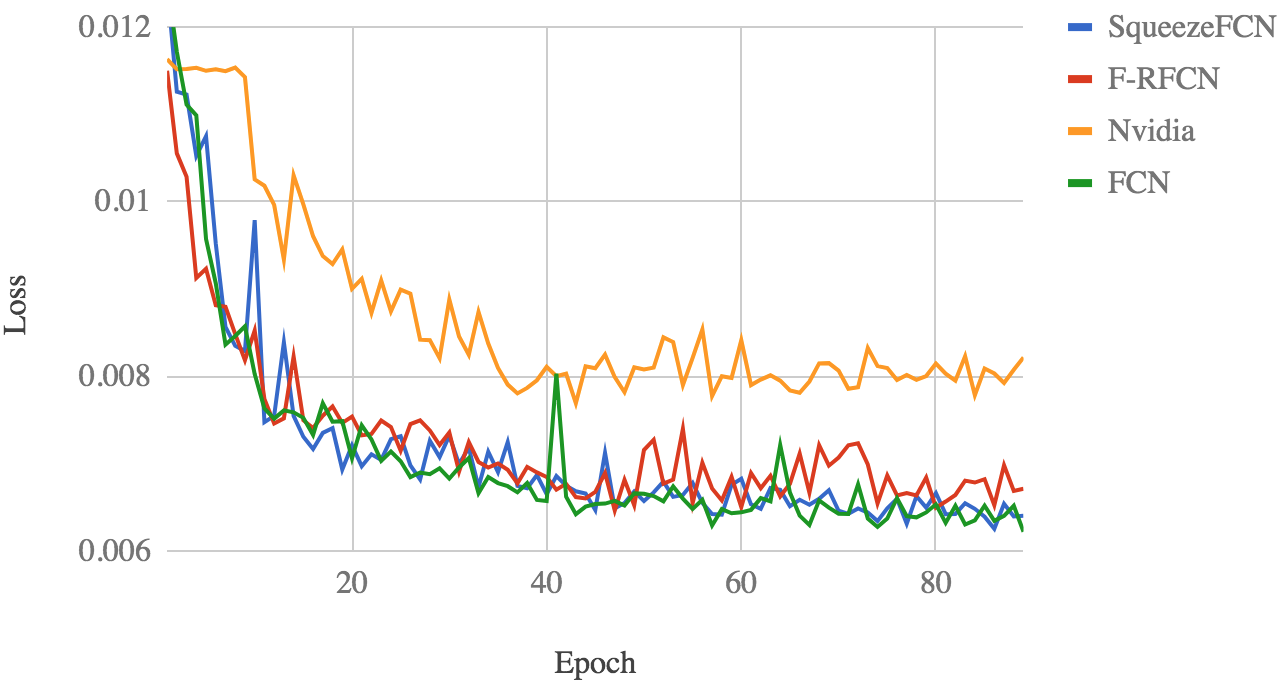}
\caption{Loss by 0.1\% subsampled epoch}
\label{fig:nonsparse-losses}
\end{center}
\end{figure}

With full dataset sampling, we mainly see the shared effect of the deep architecture. Our choice of ELU, average pooling, BatchNorm, and dropouts shares the conventional results of improved generalization and better losses. We can see the loss curves in Figure [\ref{fig:nonsparse-losses}]. To examine the differences between our three networks, we now examine the effects of sparse dataset sampling.

\subsubsection{Sparse Dataset Sampling}

\begin{table}[t]
\begin{center}
\begin{tabular}{c|ccccc}
\multicolumn{1}{c}{\bf}  &\multicolumn{1}{c}{\bf FCN} &\multicolumn{1}{c}{\bf SqueezeFCN} &\multicolumn{1}{c}{\bf F-RFCN} &\multicolumn{1}{c}{\bf Nvidia*}
\\ \hline \\
$r$         & 0.985 & 0.979 & \textbf{0.825} & 0.988\\
$r_{div}$   & 1.048 & 1.051 & \textbf{1.008} & \textbf{1.007}\\
$\sigma_r$  & 1.032 & \textbf{1.026} & 1.096 & 1.033\\
$\sigma_{r, div}$  & 1.136 & 1.120 & \textbf{1.019} & 1.037\\
$min_{\forall e}(l)$ & 0.013 & 0.009 & \textbf{0.0077} & 0.009\\
epochs & 310 & 229 & \textbf{24} & 233\\
\end{tabular}\vspace{0.2cm}
\caption{Comparison of validation loss function convergence under sparse sampling}
\label{table:1}
\end{center}
* Altered for multiple input/output time frames
\end{table}

Table [\ref{table:1}] shows further results retrieved with a sparse dataset sampling method. Firstly, we see that SqueezeFCN and the F-RFCN both either meet or beat the performance of the Nvidia model under sparse sampling. Most notably, the F-RFCN network has a significantly lower validation loss of $0.0076$ compared to Nvidia's $0.009$. Additionally, we see that the F-RFCN converges almost an entire order of magnitude faster than the Nvidia model does. This demonstrates that the feedback system described above has a significant effect on the convergence of the LSTM under sparse data conditions.\\

\begin{figure}
\begin{center}
\includegraphics[scale=0.35]{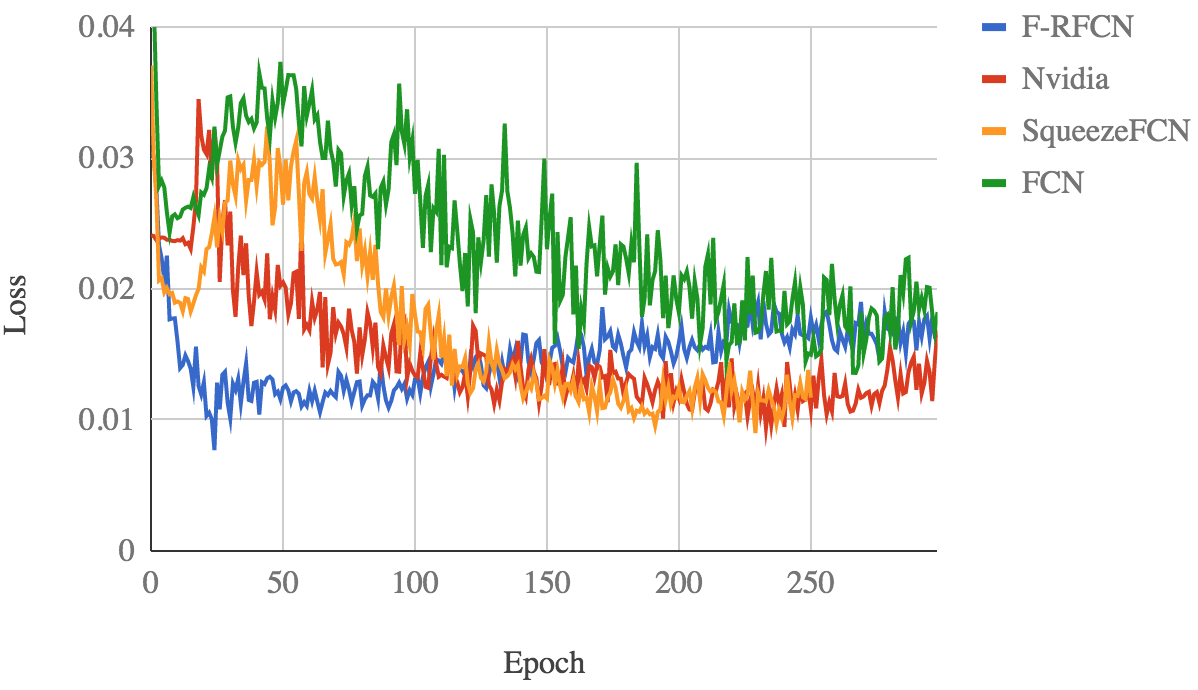}
\caption{Loss by 0.1\% subsampled epoch with sparse sampling}
\label{fig:losses}
\end{center}
\end{figure}

Meanwhile, the SqueezeFCN converges at almost the same rate, but arrives at a lower minimum validation loss than Nvidia's model, demonstrating that the inherent classification properties of the Fire module contribute significantly to the discriminative and perceptive abilities of the network. We can see the convergence effects agree with a qualitative evaluation of the loss in Figure [\ref{fig:losses}]. The FCN model, on the other hand, performs very poorly here, implying that the deeper network only works when combined with other improvements such as the Fire layer architecture and/or the LSTM architecture. Note, however, that the improvement differences of the three proposed models only appear under sparse data conditions. \\

\subsection{Experimental Improvements}

\begin{figure}
\begin{center}
\includegraphics[scale=0.7]{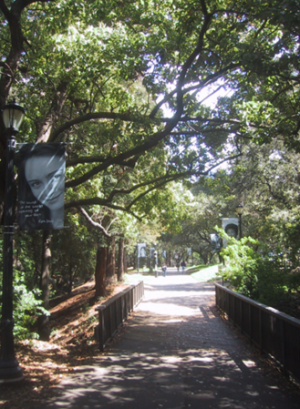}
\caption{Part of the test track}
\label{fig:test-track}
\end{center}
\end{figure}

Our models were validated on a live test track surrounded by grass and bark as shown in Figure [\ref{fig:test-track}]. The track has a few curves and no branching points so the car would always have a clear signal of where it should be going. We measure a few key benchmarks: average time to failure, average distance to failure, and \% autonomy\footnote[1]{\% autonomy is calculated by $\dfrac{t-6n}{t}$ where t is the total time in seconds and n is the number of failures in said time.} as described by Nvidia \cite{Nvidia} for two models, the F-RFCN and the modified Nvidia model. We count failure as when both front wheels run off of the test track. The model was tested on the track 10 times in each direction for both models to collect data. Because night driving was not in the dataset, the model was also tested qualitatively at night to determine the extent of generalization by the models.

\subsubsection{Test Track Results}

\begin{table}[t]
\begin{center}
\begin{tabular}{c|cc}
\multicolumn{1}{c}{\bf}  &\multicolumn{1}{c}{\bf F-RFCN} &\multicolumn{1}{c}{\bf Nvidia}
\\ \hline \\
Average time to failure (s) & 124 & 72.75\\
Average distance to failure (m) & 132.9 & 33.4\\
\% Autonomy & 85\% & 82\%\\
\end{tabular}\vspace{0.2cm}
\caption{Experimental results}
\label{table:experimental}
\end{center}
\end{table}

The Nvidia and F-RFCN models both navigate the curves and are able to show path following capabilities up to at least one minute. However, the F-RFCN clearly performs better with respect to all of our metrics. Since the network models were also given control over the motor, it has to be noted that the Nvidia model drove slower and therefore average time to failure and distance to failure should be compared jointly. As Table [\ref{table:experimental}] illustrates, the Nvidia model consistently fails after less distance than the F-RFCN model does. Note that the average distance to failure for the F-RFCN model is over 4x further than the Nvidia model's. In addition, the time to failure for the F-RFCN model is much higher, with it driving for 70\% more time than the Nvidia model does before failing. Finally, we can see that the F-RFCN has a 3\% autonomy gain over Nvidia's model.

\subsubsection{Qualitative Results}

The F-RFCN network was able to generalize much more than the Nvidia model could. When testing at night, the Nvidia model could not sense curves very accurately and would consistently fail to detect the edge of the path. Even after running off of the path, the model could not extract enough information to steer. On the other hand, the F-RFCN network also failed often, but it would attempt to steer when it got very close to the edge of the path. Even after veering off of the path, it continued to attempt to correct itself, eventually arriving back on the track.\\

We also tested the models by shining flashlights at the vehicle during the night. The Nvidia model was attracted to light during the night, and while the F-RFCN network initially followed the light, once it approached a detectable edge it once again began to attempt to correct its own path.

\section{Conclusion}
In conclusion, we designed three different network models which solve the end-to-end driving task in order to explore alternative design rationals. They were created as three samples in the space of network designs, comparing fully-convoluted, compression-based, and spatiotemporal input processing to predict steering and motor commands. %
We showed that we are able to outperform the baseline network in terms of minimum loss function value, convergence time, and parameter count, all while achieving similar stability of convergence. 
These three network designs show an improved loss compared with the baseline, and their different convergence behavior is further shown through the results from using a sparse sampling method.\\%

Experimentally, we showed that the F-RFCN network outperforms the baseline network in terms of time to failure, distance to failure, and percent autonomy. Additionally, the F-RFCN network qualitatively shows better generalization over environments than the Nvidia model does. 
Our results emphasize the importance of network design paradigms such as BatchNorm, multi-task learning, and deep and thin networks. Additionally, we demonstrated the boosting effects in training of increased gradient feedback and semantic encoding and decoding through LSTMs.

\section{Acknowledgements}
We thank Berkeley Deep Drive, the International Computer Science Institute, and the University of California, Berkeley, for sponsorship of the group and project.

We also thank Jordan Hart (jordanhart@berkeley.edu) for assisting in the setup of the model vehicles, experimentation, and paper review. 

{\small
\bibliographystyle{ieee}
\bibliography{egbib}
}
\end{document}